\documentclass[letterpaper]{article}
\usepackage{aaai2026}
\usepackage{times}
\usepackage{helvet}
\usepackage{courier}
\usepackage[hyphens]{url}
\usepackage{graphicx}
\usepackage{subcaption}

\usepackage{natbib}
\usepackage{caption}
\usepackage[linesnumbered,ruled,vlined]{algorithm2e}
\usepackage{listings}
\usepackage{booktabs,makecell}
\usepackage{amsmath}

\title{SIGN: Schema-Induced Games for Naming}

\author{Ryan Zhang\textsuperscript{\rm 1} \quad Herbert Woisetschläger\textsuperscript{\rm 2}}

\affiliations{
\textsuperscript{\rm 1}Horace Greeley High School, \\
\textsuperscript{\rm 2}Technical University of Munich, Munich, Germany \\
ryzhangofficial@gmail.com, h.woisetschlaeger@tum.de
}

\begin{document}
\maketitle

\begin{abstract}
Real-world AI systems are tackling increasingly complex problems, often through interactions among Large Language Model (LLM) agents. When these agents develop inconsistent conventions, coordination can break down. Applications such as collaborative coding and distributed planning therefore require reliable, consistent communication, and scalability is a central concern as systems grow. We introduce \textbf{S}chema-\textbf{I}nduced \textbf{G}ames for \textbf{N}aming (SIGN), a naming game that examines how lightweight structure can steer convention formation. We compare schema-induced communication to unconstrained natural language and find faster convergence with up to $5.8\times$ higher agreement. These results suggest that minimal structure can act as a simple control knob for efficient multi-agent coordination, pointing toward broader applications beyond the naming game.
\end{abstract}

\section{Introduction}
Large language models (LLMs) are central to applications such as chat assistance, code completion, and summarization, but are typically studied in isolation. Recent work has begun to explore multi-agent settings and simulated societies, where agents coordinate and form conventions~\cite{guo2024large}. Naming-game studies show that shared conventions can emerge from interaction alone~\cite{ashery2025emergent}, and even with limited memory a population can converge to a common naming scheme, much like linguistic convention evolution in human groups. In parallel, LLMs can be instructed to use structured formats such as JSON schemas or templates, which improve reasoning and collaboration while reducing verbosity~\cite{chen2024beyond}. Despite evidence for emergent conventions and the benefits of structure, it remains unclear whether lightweight schema priors can \emph{steer} convention formation itself.

As LLM multi-agent systems research grows, developing simple, efficient, and controllable protocols is becoming increasingly relevant. Enforcing a schema provides a straightforward, model-agnostic control knob. This study tests whether imposing a minimal message schema in a population naming game~\cite{baronchelli2008naming} will (i) reduce tokens-to-convergence and (ii) improve overall population agreement.

\section{Methodology}
The naming game is defined over a population of $N$ agents and a fixed lexicon $\mathcal{L}={C_1,\dots,C_M}$. Time advances in rounds $t=1,\dots,T$, with agents paired uniformly at random. Agent $i$ produces a message $m_i^t$, which a decoder maps to a name $y_i^t \in \mathcal{L}$. Each agent keeps a memory window of size $K$ containing its last $K$ partner-only interactions, which guides future proposals. We study three conditions: Natural Language (NL), Natural Language Sliding Window (NL-SW), and Schema.

In NL, agents generate unconstrained natural language outputs, and the decoder extracts a valid token if possible. NL-SW extends this by incorporating memory $K$, so recent interactions influence each proposal.

\begin{algorithm}[h]
\DontPrintSemicolon
\SetAlgoLined
\KwIn{$N$, $\mathcal{L}$, $K$, $T$, $\alpha$}
\For{$t=1$ \KwTo $T$}{
  Pair agents $i,j$ uniformly at random\;
  Each forms proposal $m^t$ using partner-only $K$\;
  Parse \texttt{@say \{name: Ck\}} $\rightarrow y$\;
  \If{non-compliant}{
    Retry once with reminder\;
    \If{still invalid}{decode free text; \If{undecodable}{$y \leftarrow$ None}}
  }
  \If{$y_i \ne y_j$}{adopt partner’s $C_k$ w.p.~$\alpha$}
}
\caption{Schema-Induced Condition}\label{alg:schema}
\end{algorithm}

\begin{figure*}[t]
\centering
\includegraphics[width=.24\textwidth]{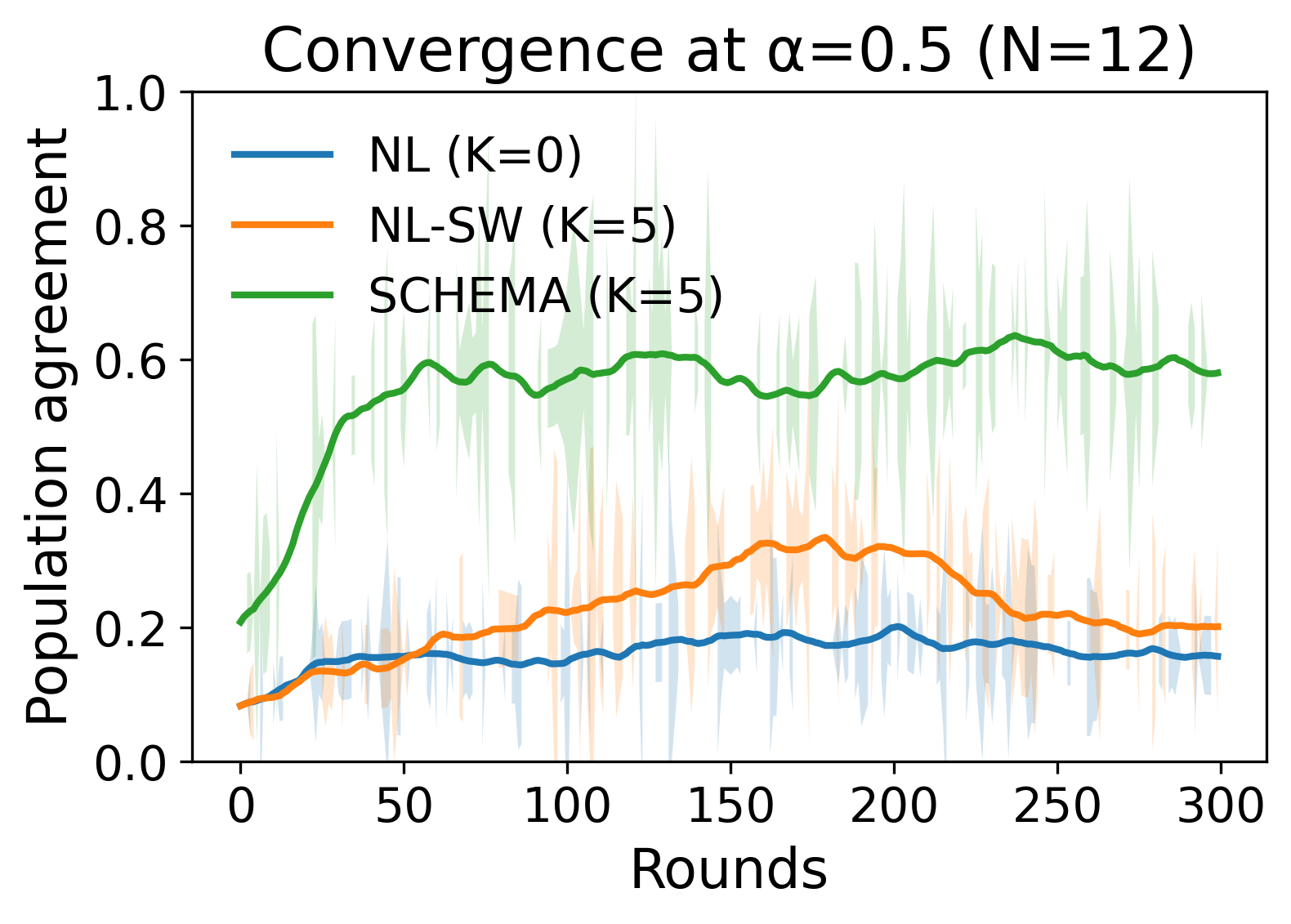}\hfill
\includegraphics[width=.24\textwidth]{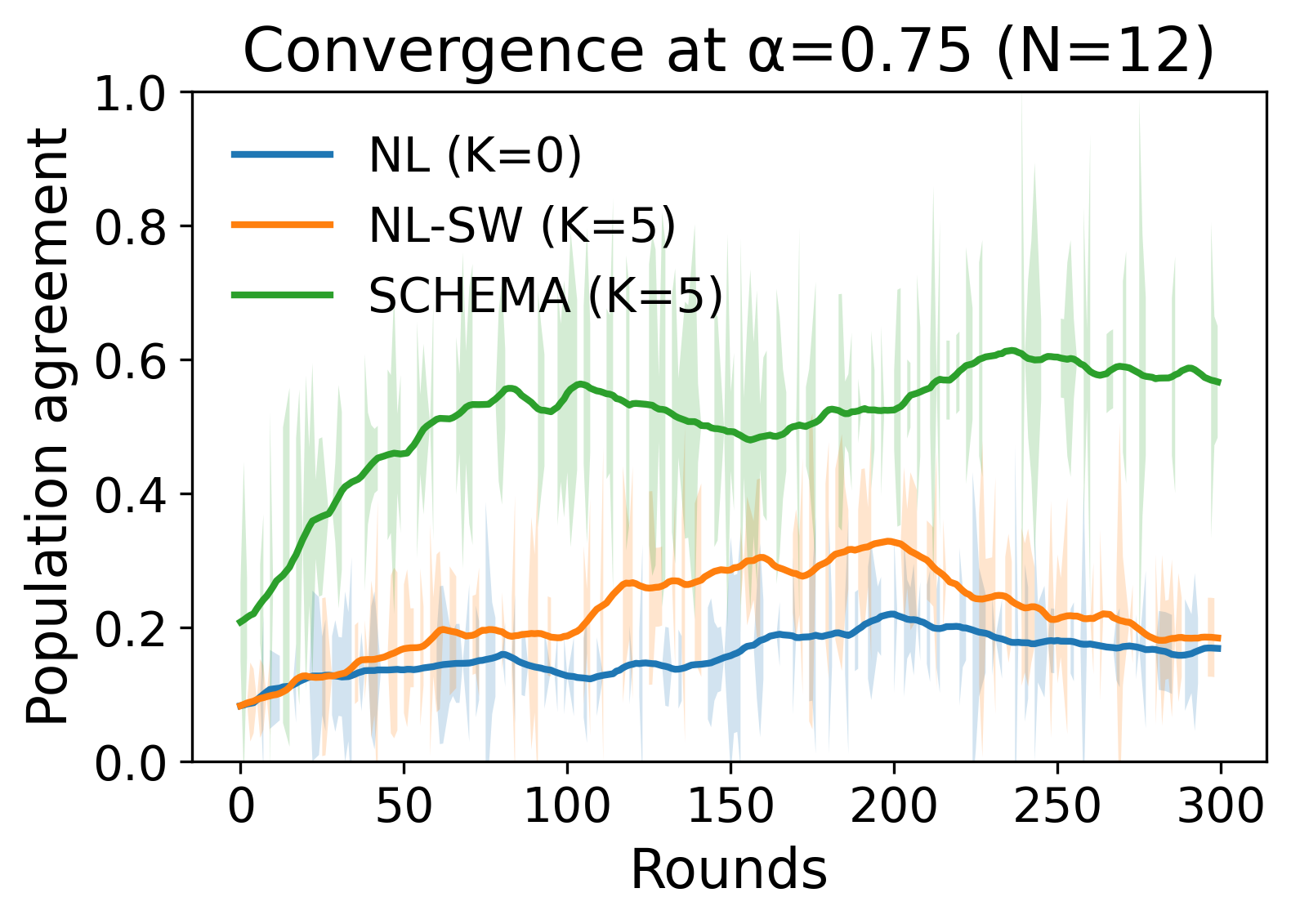}\hfill
\includegraphics[width=.24\textwidth]{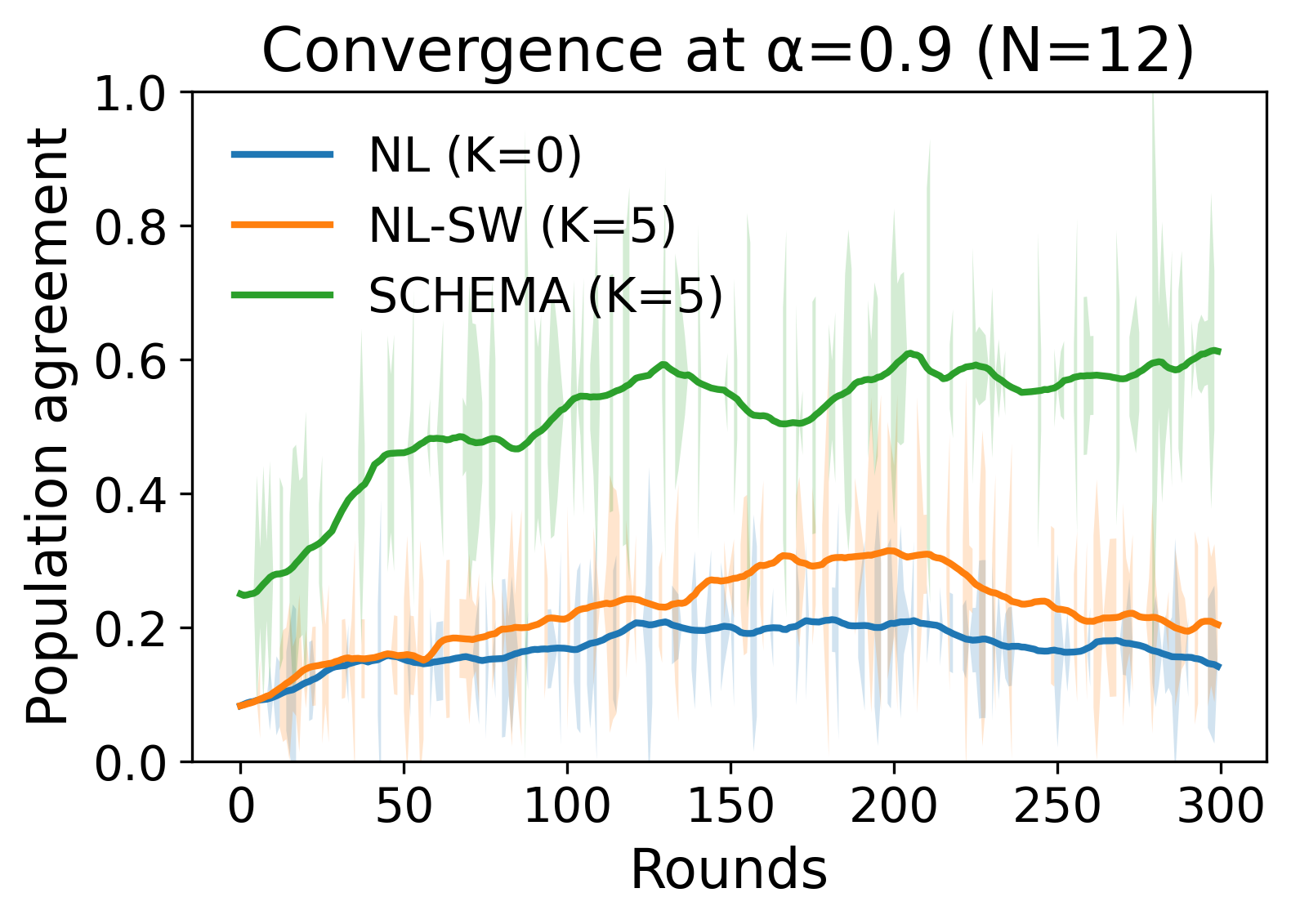}\hfill
\includegraphics[width=.24\textwidth]{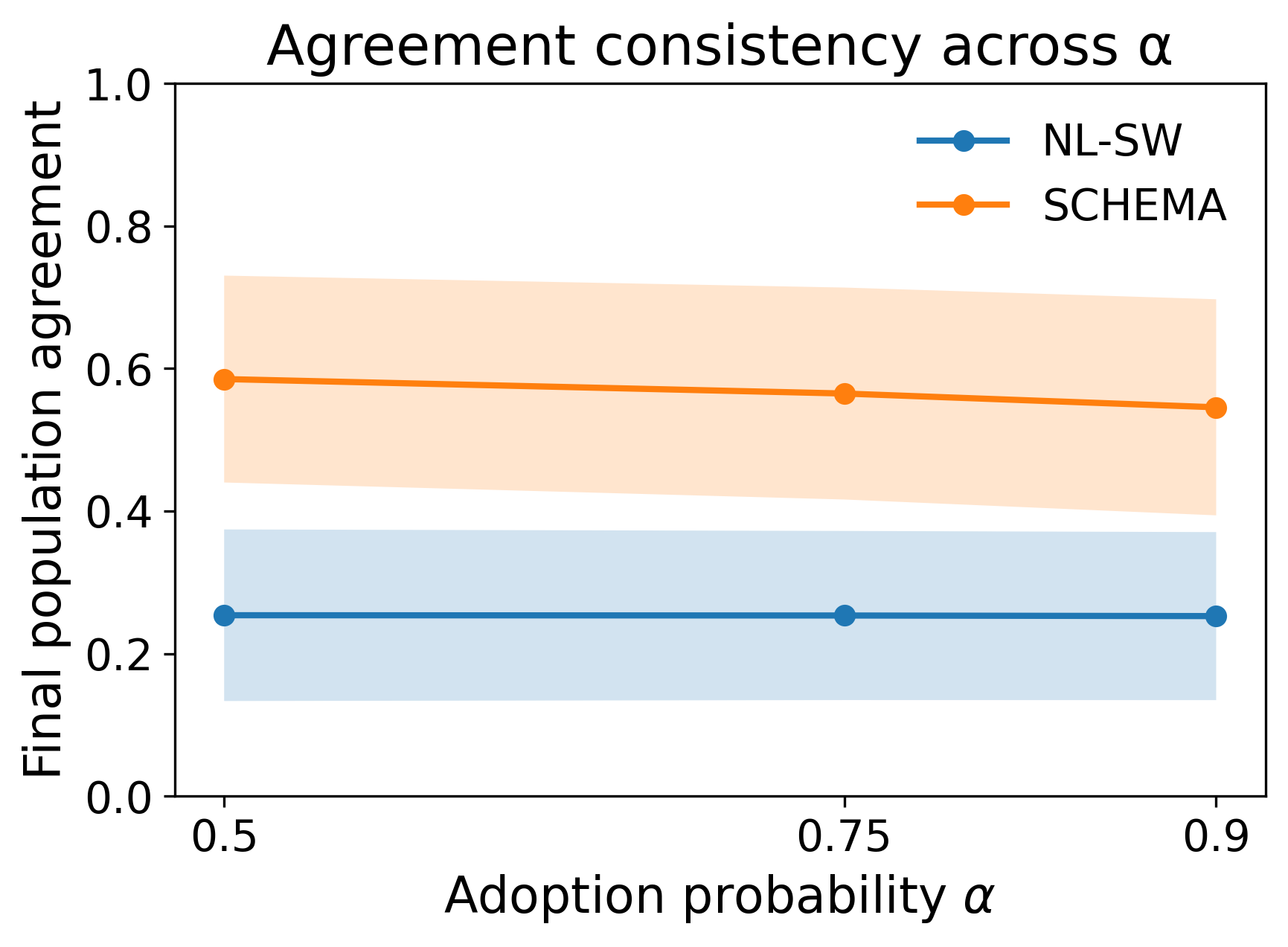}

\vspace{0.25em}

\makebox[.24\textwidth]{\scriptsize $\alpha=0.5$}\hfill
\makebox[.24\textwidth]{\scriptsize $\alpha=0.75$}\hfill
\makebox[.24\textwidth]{\scriptsize $\alpha=0.9$}\hfill
\makebox[.24\textwidth]{\scriptsize Final agreement}

\caption{Population agreement under different adoption probabilities $\alpha$ with $N=12$ and $K=5$.}
\label{fig:convergence_alpha}
\end{figure*}

The Schema condition is outlined in Algorithm~\ref{alg:schema} and requires replies to match \verb|@say {name: Ck}|, from which a regex parser extracts the $C_k$ token. The lightweight schema tag follows earlier work showing that structured templates improve reasoning and reliability in LLM tasks~\cite{chen2024beyond}. This design provides agents an explicit, easily parsed handle for lexicon entries, keeping replies transparent to the listener with minimal overhead. Non-compliant outputs receive one retry with a brief reminder. If still invalid, the output defaults to a random valid $C_k$ and is marked non-compliant. The variable $\alpha$ denotes the lose shift probability, meaning that after a mismatch an agent adopts the partner’s decoded name with probability $\alpha$.

\section{Experiments}
We evaluate the naming game with populations of $N \in \{12,24\}$ agents, lexicon size $M{=}12$, and $T{=}300$ rounds, defining tokens-to-convergence as the number of tokens needed for the population to reach a chosen agreement. Memory windows vary with $K \in \{5,10\}$ and lose-shift with $\alpha \in \{0.5,0.75,0.99\}$. Each configuration is run with three random seeds. Agents use the Phi-3 Mini 4K Instruct model with fixed decoding: max new tokens $=32$, temperature $=0.7$, top-$p=0.9$, and repeat penalty $=1.1$.

\begin{table}[t]
\caption{Average population agreement by condition with varying agents ($N$) and memory ($K$).}
\label{tab:population_agreement}
\centering
\small
\setlength{\tabcolsep}{6pt}
\begin{tabular}{@{}ccccc@{}}\toprule
\multicolumn{2}{c}{} & \multicolumn{3}{c}{\textbf{Population Agreement}} \\
\cmidrule(lr){3-5}
N & K & NL & NL-SW & SCHEMA \\
\midrule
12 & 0  & 0.111 {\scriptsize$\pm$ 0.048} & \textemdash & \textemdash \\
24 & 0  & 0.125 {\scriptsize$\pm$ 0.042} & \textemdash & \textemdash \\
12 & 5  & \textemdash & 0.278 {\scriptsize$\pm$ 0.127} & \textbf{0.611} {\scriptsize$\pm$ 0.293} \\
24 & 5  & \textemdash & 0.292 {\scriptsize$\pm$ 0.042} & \textbf{0.556} {\scriptsize$\pm$ 0.064} \\
12 & 10 & \textemdash & 0.333 {\scriptsize$\pm$ 0.144} & \textbf{0.639} {\scriptsize$\pm$ 0.096} \\
24 & 10 & \textemdash & 0.295 {\scriptsize$\pm$ 0.039} & \textbf{0.588} {\scriptsize$\pm$ 0.085} \\
\bottomrule
\end{tabular}
\end{table}

\begin{figure}[t]
\centering
\includegraphics[width=0.75\linewidth,height=0.20\textheight]{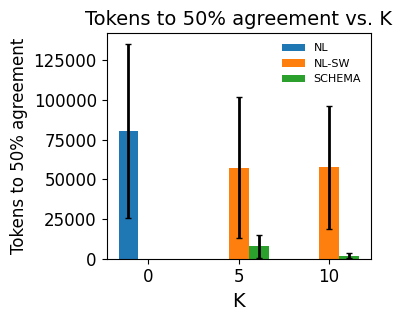}
\caption{Tokens required to reach 50\% population agreement across conditions.}
\label{fig:tokens_to_conv}
\end{figure}

As shown in Figure~\ref{fig:convergence_alpha}, the Schema condition achieves substantially higher population agreement than both NL and NL-SW across adoption probabilities $\alpha$. Agreement under Schema rises toward $0.6$–$0.65$, while NL-SW peaks near $0.3$ and NL remains below $0.2$. Increasing $\alpha$ slightly lowers agreement for both NL-SW and Schema. Table~\ref{tab:population_agreement} reports average agreement across populations and memory sizes. Agreement is stable for Schema, with decreases from $N=12$ to $N=24$ and modest increases from $K=5$ to $K=10$, indicating that gains come mainly from schema induction rather than population size or memory. Beyond higher agreement, standard deviation decreases over time, with Schema at $\alpha=0.5$ producing the most consistent outcomes. Figure~\ref{fig:tokens_to_conv} shows that Schema converges with an order of magnitude fewer tokens than NL or NL-SW.

\section{Conclusion and Future Work}
Inspired by work showing that conventions can emerge from interaction and by the use of structured formats in supervised tasks, we show that adding a fixed schema to LLM agents steers convention formation in a naming game, yielding up to $5.8\times$ greater population agreement. Minimal structural priors thus can shape how conventions emerge. A key direction is testing whether schema reduces variation in LLM responses and whether this consistency may limit broader tasks, together with studies of larger populations, varied lexicon sizes, and alternative schema designs.

\bibliography{references}

\begin{thebibliography}{4}
\providecommand{\natexlab}[1]{#1}

\bibitem[{Ashery, Aiello, and Baronchelli(2025)}]{ashery2025emergent}
Ashery, A.~F.; Aiello, L.~M.; and Baronchelli, A. 2025.
\newblock Emergent social conventions and collective bias in LLM populations.
\newblock \emph{Science Advances}, 11(20): eadu9368.

\bibitem[{Baronchelli, Loreto, and Steels(2008)}]{baronchelli2008naming}
Baronchelli, A.; Loreto, V.; and Steels, L. 2008.
\newblock In-depth analysis of the Naming Game dynamics: the homogeneous mixing case.
\newblock \emph{arXiv:0803.0398}.

\bibitem[{Chen et~al.(2024)Chen, Yuan, Yuan, Su, Qian, Yang, Xie, Liu, and Sun}]{chen2024beyond}
Chen, W.; Yuan, C.; Yuan, J.; Su, Y.; Qian, C.; Yang, C.; Xie, R.; Liu, Z.; and Sun, M. 2024.
\newblock Beyond natural language: Llms leveraging alternative formats for enhanced reasoning and communication.
\newblock \emph{arXiv preprint arXiv:2402.18439}.

\bibitem[{Guo et~al.(2024)Guo, Chen, Wang, Chang, Pei, Chawla, Wiest, and Zhang}]{guo2024large}
Guo, T.; Chen, X.; Wang, Y.; Chang, R.; Pei, S.; Chawla, N.~V.; Wiest, O.; and Zhang, X. 2024.
\newblock Large language model based multi-agents: A survey of progress and challenges.
\newblock \emph{arXiv preprint arXiv:2402.01680}.

\end{thebibliography}

\clearpage
\section{Appendix} 
Here we provide additional experimental results not reported in the main findings. 

\subsection{Code Availability} 
Code to reproduce all experiments is available at \url{https://github.com/ryanzhangofficial/llm-naming-game-steering}. 

\subsection{LLaMA-Only Experiments} 
We ran the Llama 3.2 3B Instruct model with the same decoding settings as Phi-3 Mini 4K Instruct: max new tokens $=32$, temperature $=0.7$, top-$p=0.9$, and repeat penalty $=1.1$. Each configuration used three random seeds with $N=12$, $K=5$, and $\alpha=0.75$. 

\begin{figure}[h] 
\centering 
\includegraphics[width=0.8\linewidth]{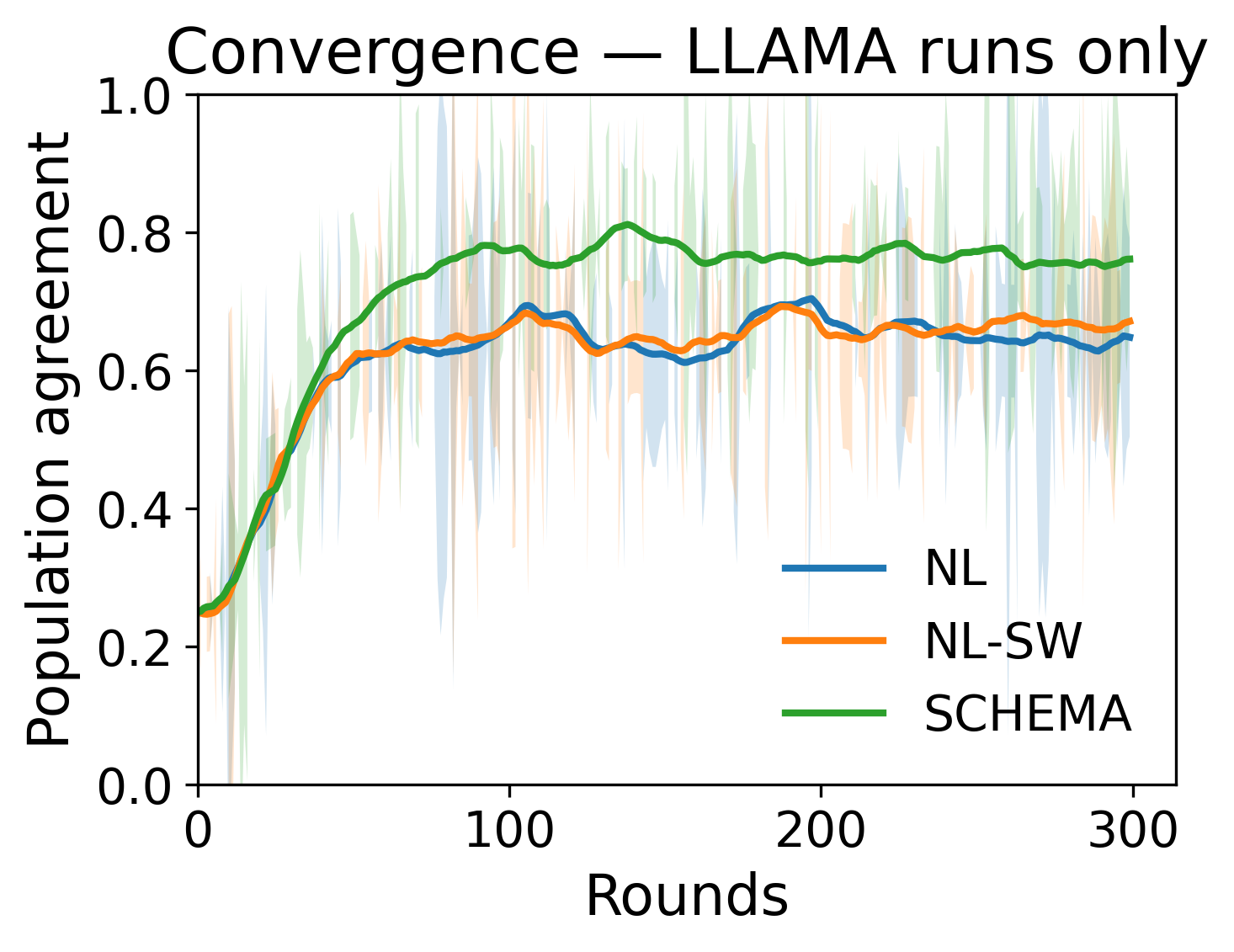} 
\caption{Convergence of the naming game using Llama 3.2 3B Instruct.} 
\label{fig:convergence_llama_only} 
\end{figure} 

Figure~\ref{fig:convergence_llama_only} shows that Schema also improves agreement with only LLaMA models. Agreement rises toward $0.75$–$0.8$ under Schema, compared to $0.65$–$0.7$ for NL and NL-SW. Overall agreement exceeds that of the Phi-only runs, despite comparable model size, suggesting that model family differences shape naming game dynamics, while Schema continues to provide an advantage.

\subsection{Mixed Model Experiments} 
We used a mixed population of 12 agents, with 6 Phi-3 Mini 4K and 6 Llama 3.2 3B Instruct models. Decoding settings matched earlier experiments. Due to time limits, these runs were capped at 100 rounds. 

\begin{figure}[h] 
\centering 
\includegraphics[width=0.8\linewidth]{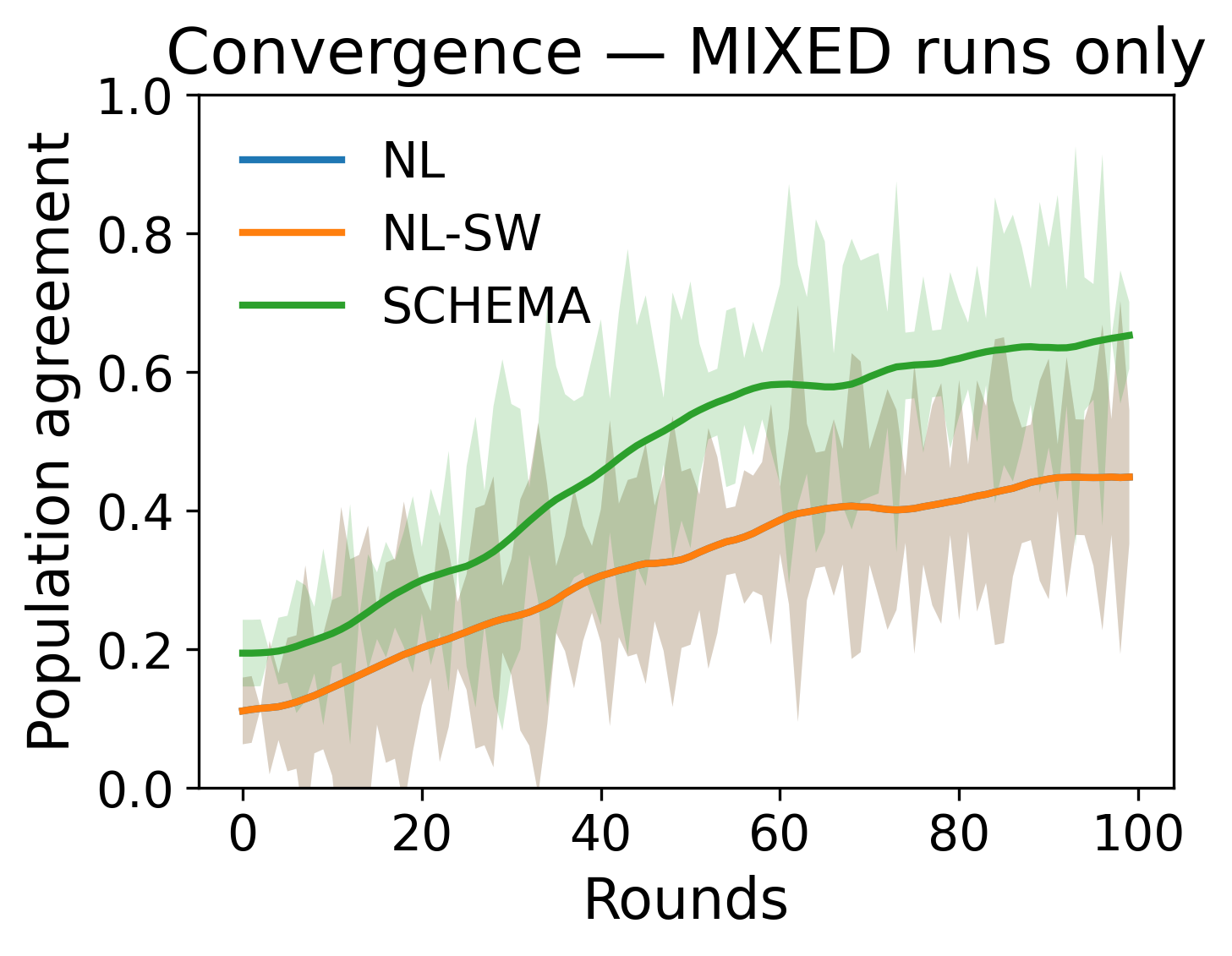} 
\caption{Convergence of the naming game using a mixed population of Phi and LLaMA models.} 
\label{fig:convergence_mixed} 
\end{figure} 

Figure~\ref{fig:convergence_mixed} shows that Schema again provides a clear advantage, achieving higher agreement than NL and NL-SW despite the heterogeneous population. Future work will extend these experiments to longer runs and to mixes of more than two model families. 

\subsection{Tokens-to-Convergence at Higher Thresholds} 
\begin{figure}[!ht]
\centering

\begin{subfigure}{0.82\linewidth}
\centering
\includegraphics[width=\linewidth]{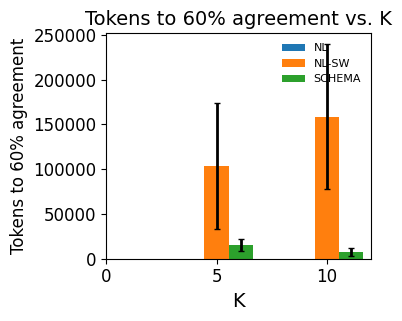}
\caption{Tokens required to reach $60\%$ agreement.}
\label{fig:tokens_to_conv_06}
\end{subfigure}

\vspace{6pt}

\begin{subfigure}{0.82\linewidth}
\centering
\includegraphics[width=\linewidth]{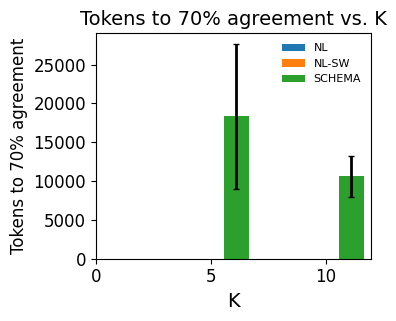}
\caption{Tokens required to reach $70\%$ agreement.}
\label{fig:tokens_to_conv_07}
\end{subfigure}

\caption{Tokens-to-convergence results at higher thresholds of agreement.}
\label{fig:tokens_to_conv_higher}
\end{figure}

In the main text, Figure~\ref{fig:tokens_to_conv} reported tokens required to reach $50\%$ agreement, where Schema converged with an order of magnitude fewer tokens than NL or NL-SW. Figures~\ref{fig:tokens_to_conv_higher}\subref{fig:tokens_to_conv_06} and \ref{fig:tokens_to_conv_higher}\subref{fig:tokens_to_conv_07} extend this to higher thresholds. At $60\%$ agreement, Schema converges with nearly two orders of magnitude fewer tokens than NL-SW, while NL never reaches the threshold. At $70\%$, only Schema achieves convergence, requiring slightly more tokens than at $60\%$. These results demonstrate that schema induction enables convergence even under strict criteria.

\end{document}